\documentclass{llncs}

\usepackage[utf8]{inputenc} % allow utf-8 input
\usepackage[T1]{fontenc}    % use 8-bit T1 fonts
\usepackage{hyperref}       % hyperlinks
\usepackage{url}            % simple URL typesetting
\usepackage{booktabs}       % professional-quality tables
\usepackage{amsfonts}       % blackboard math symbols
\usepackage{amsmath}       % blackboard math symbols
\usepackage{nicefrac}       % compact symbols for 1/2, etc.
\usepackage{microtype}      % microtypography
\usepackage{graphicx}
\usepackage{wrapfig}

\title{Large-Scale Shape Retrieval with Sparse 3D Convolutional Neural Networks}

\author{Alexandr Notchenko\inst{1}\inst{2} \and Yermek Kapushev\inst{1}\inst{2}
\and Evgeny Burnaev\inst{1}
}

\institute{
Skolkovo Institute of Science and Technology,\\
\email{\{alexandr.notchenko, yermek.kapushev\}@skolkovotech.ru} \\
\email{e.burnaev@skoltech.ru}
\and
Institute of Information Transmission Problems}

\begin{document}
% \nipsfinalcopy is no longer used

\maketitle

\begin{abstract}
  In this paper we present results of performance evaluation of S3DCNN --- a Sparse 3D Convolutional Neural Network --- on a large-scale 3D Shape benchmark ModelNet40, and measure how it is impacted by voxel resolution of input shape. We demonstrate comparable classification and retrieval performance to state-of-the-art models, but with much less computational costs in training and inference phases. We also notice that benefits of higher input resolution can be limited by an ability of a neural network to generalize high level features.
\end{abstract}

\section{Introduction}

For computer vision systems a precise and robust operations in real environments is only possible by harnessing information from 3D data. To achieve this we need to overcome some challenges of this kind of problems.

Data, received from such devices as $2.5D$ scanners, is often given in the form of noisy meshes or point clouds, which is not the best fit for new kinds of models such as Convolutional Neural Networks (CNN's) \cite{lecun1995convolutional}.

In the current state-of-the-art systems Convolutional Neural Networks are widely used, their effectiveness at processing 2D images is also suggestive of their efficacy to process 3D objects if presented in the form of several rendered views of the object. For example on one ModelNet40 \cite{wu20153d} benchmark three recent papers, based on this idea, showed incremental improvements in recognition performance \cite{su15mvcnn,johns2016pairwise,hegde2016fusionnet}. However, it can be argued that high performance is predicated by the usage of CNNs pre-trained on ImageNet \cite{deng2009imagenet}.

Voxel representation of 3D shapes (i.e. a shape is represented as a three-dimensional grid, where occupied cells are binary values) are compatible with ConvNets input layers but create a number of difficulties.
Adding a third spatial dimension in the input grid correspondingly increases computational costs. Number of cells scales as a power of three w.r.t. the resolution of the voxel grid. Low resolution grids make it difficult to differentiate between similar shapes, and lose some of the fine details available in 2D renderings of equivalent resolution.

Some 3D Dense Convolutional Networks have been evaluated on the ModelNet40 benchmark \cite{maturana2015voxnet,sedaghat2016orientation,wu20153d,brock2016generative}, but they still not perform as well as their multi-rendering 2D counterparts.

At the same time using Modified Spatially Sparse Neural Networks algorithms \cite{graham2014spatially} to process data we are able to have reasonable training and inference time even with input resolution up to $100^3$ voxels.

In this work, we present Sparse 3D Deep Convolutional Neural Networks and explore their ability to perform large-scale shape retrieval on the popular benchmark ModelNet40 \cite{wu20153d} depending on an input resolution and a network architecture.
% and ShapeNetCore55 \cite{chang2015shapenet},

Sparse 3D CNNs are able to generate relevant features for retrieval analogously to 2D extractors. To have a system that uses many 2D rendered projections for inference is computationally very costly, especially for the task of Large Scale 3D Shape Retrieval. In this paper we present some preliminary results of our attempt to find out if the resolution of an object Voxelization impacts on descriptive feature extraction as measured by the retrieval performance on a sufficiently big dataset. Also we demonstrate ability of Sparse 3D CNNs to perform metric learning in the triplet loss setup. Lastly we train our model to perform classification on the ModelNet40 benchmark.

In Section~\ref{sec:2} we formulate the problem in more detail and discuss latest relevant methods.
In Section~\ref{sec:3} we describe our approach to neural networks that helps us to solve the problem posed in Section~\ref{sec:2}.
In Section~\ref{sec:4} we document conditions of computational experiments we performed.
In Section~\ref{sec:5} we discuss results and make conclusions about our approach to the problem.

% Performing triplet metric learning with 3D CNNs was done for ranking architectural 3D models by style\cite{Lim:2016:StyleLearning}.

% \subsection{Related work}

\section{3D Large-Scale Retrieval}
\label{sec:2}
\subsection{Large-Scale 3D Shape datasets}

As can be seen the great improvements in recent years for the problem of 2D large-scale image recognition, are not just the result of wide-spread adoption of Deep Learning techniques, but also it is due to the availability of large datasets that capture sufficient variety of features at different scales to be representative of some domain.
However, only recently in the 3D recognition and retrieval such datasets started being published.

The recent competition ModelNet evaluated several models utilizing Neural Networks for 3D retrieval. ModelNet40 is a subset of this dataset, and it is going to be our main benchmark for the retrieval task.
The approach for creating descriptors from multiple projections of a 3D shape with a transfer learning from ImageNet showed the best performance \cite{su15mvcnn}. No full 3D algorithms that process voxels directy have been described up to now.

\subsection{Shape descriptors}

To make inferences about 3D objects for purposes of computer vision or computer graphics, researchers developed a big amount of shape descriptors\cite{kazhdan2003rotation,knopp2010hough,bronstein2011shape,kokkinos2012intrinsic}.

Shape descriptors usually fit into two categories: one where shape descriptors are computed using 3D representations of objects, e.g. voxel discretizations, meshes, point clouds, or implicit surfaces, and the second one that describes a shape of a 3D object by a collection of 2D projections, often from multiple viewpoints.

Before large-scale 3D shape datasets such as ModelNet \cite{wu20153d} and 3dShapeNet model which learns shape descriptors from voxel representation of a mesh object through 3D convolutional nets, 3D shape descriptors were mostly special functions capturing specific geometric properties of the shape surface or volume, for example: spherical functions computed on volumetric grids \cite{kazhdan2003rotation}, generalization of SIFT and SURF feature descriptors for voxel grids \cite{knopp2010hough}, or for non-rigid bodies and deformable shapes heat kernel signatures on meshes \cite{bronstein2011shape,kokkinos2012intrinsic}. Developing classifiers and other supervised machine learning algorithms on top of such 3D shape descriptors poses a number of challenges. The success of CNNs image descriptors allows us to hope that descriptors based on 3D convolutional nets can be also beneficial compared to classic descriptors.

\subsection{Triplet learning}
Recent work in \cite{hoffer2015deep} shows that learning representations with triplets of examples
gives much better results than learning with pairs using the same network. Inspired by this, we
focus on learning feature descriptors based on triplets of patches.

Learning with triplets involves training from samples of the form $(a, p, n)$, where 
\begin{itemize}
\item  $a$ is an anchor object,
\item $p$ denotes a positive object, which is a sample we want to be closer to $a$ and usually being a different sample of the same class as $p$, and
\item $n$ is a negative sample belonging to a different class than $a$ and $p$.
\end{itemize}
Optimizing parameters of the network brings $a$ and $p$ close in the feature space, and pushes apart $a$ and $n$.

Finally, let us introduce this triplet loss, also known as the ranking loss. It was first proposed for learning embedding using CNNs in \cite{wang2014learning} and can be defined as follows:
\begin{itemize}
\item Let us define $\delta_+ = \mathrm{cosine}(f(a), f(p))$ and $\delta_- = \mathrm{cosine}(f(a), f(n))$, i.e. this is a cosine distance between some feature representations $f(\cdot)$ for different objects,
\item Then for a particular triplet we calculate the triplet loss using the formula
\[
\lambda ( \delta_+, \delta_- ) = \max (0, \mu + \delta_+ - \delta_- ) \,,
\]
where $\mu$ is a margin parameter. The correct order should be $\delta_- > \delta_+ + \mu$,
\item If order of objects, provided by their corresponding descriptors are incorrect w.r.t. the triplet loss, then the network adjusts its weights through back-propagation signal to reduce the error.
\end{itemize}

\section{Sparse Neural Networks}
\label{sec:3}
Using sparsity to make a neural network computations more efficient is pioneered by Benjamin Graham \cite{graham2014spatially}, who developed a low-level C++/CUDA library SparseConvNet\footnote{\url{https://github.com/btgraham/SparseConvNet}} that implements strided convolutions and max-pooling operations on a $D$-dimensional sparse tensors using GPU. Due to this inherited sparsity we are able to process data in reasonable training and inference time even with input resolution up to hundreds of voxels.
More precisely an information about voxels in a given layer is not stored in a 3-dimensional array, but in a sparse vector with active cells as elements.

Transformation of data between layers (e.g. convolutions, pooling, nonlinear activation functions), are performed on those sparse vectors. Data in areas with inactive voxels, which are most of them, does not depend on a voxel relative position, therefore it can be replaced by vectors of a smaller size without explicit spatial dimensions.

It's well known that, operating with a sparse data structures is more efficient than working with dense data.
Another useful property is that we need to store much less data for each object.
We have computed sparsity for all classes of ModelNet40 train dataset at voxel resolution equal to 40, and it's only 5.5\%.

Paper \cite{wu20153d} describes using 3D convolutions for their deep model.
Voxel labeled as active when it's intersects with a mesh object, and inactive otherwise.
This binary representation of 3D shape given as input to a 3D CNN, which has a structure similar to a 2D one.
The main problem of this approach is ineffectiveness with which data is represented and processed.
Mentioned model uses $30^3$ cells, which is approximately the number of pixels in 2D applications of CNN.
If we take into account linear dimensions it's obviously not a lot, as can be seen from Figure~\ref{fig:voxels-examples}.
That resolution was primarily chosen because of computational resource limitation.
Besides that, --- convolution is very computationally expensive operation, complexity of which rises very fast with input scale.
Computational complexity of 3D convolution for image with dimensions of $N \times M \times K$ with filters sizes of $n \times m \times k$ is equal to $\mathcal{O}(NMKnmk)$.
If we use Fast Fourier Transform (FFT), complexity can be reduced to  $\mathcal{O}((N + n)(M + m)(K + k)\log((N + n)(M + m)(K + k)))$
in exchange for more memory cost \cite{mathieu2013fast}.
But even in that case, complexity of convolutions makes it impossible to work with objects in big voxel resolutions.

\subsection{PySparseConvNet}

The SparseConvNet Library is written in C++ programming language, and utilizes a lot of CUDA capabilities for speed and efficiency. But it is very limited when it comes to 
\begin{itemize}
\item extending functionality --- class structure and CUDA kernels are very complex, and require re-compilation on every modification,
\item changing loss functions --- the only learning configuration was SoftMax with log-likelihood loss function,
\item fine grained access to layer activations --- there was no way to extract activations and therefore features from hidden layers,
\item interactivity for models exploration --- every experiment had to be a compiled binary with no way to perform operations step by step, to explore properties of models.
\end{itemize}

Because of all these problems we developed PySparseConvNet\footnote{\url{https://github.com/gangiman/PySparseConvNet}}.
On implementation level it's a python compiled module that can be used by Python interpreter, and harness all of it's powerful features. Most of modern Deep Learning tools, such as \cite{2016arXiv160502688short,tensorflow2015-whitepaper,tokui2015chainer}, use Python as a way to perform interactive computing.

Interface of PySparseConvNet is much simpler, and consist's of 4 classes:

\begin{itemize}
\item \textbf{SparseNetwork} --- Network object class, it has all the methods to change it's structure, manipulate weights and activations,
\item \textbf{SparseDataset} --- Container class for sparse samples and their labels,
\item \textbf{SparseBatch} --- Gives access to data in dataset when processing separate mini-batches,
\item \textbf{Off3DPicture} --- Wrapper class for 3D models in OFF (Object File Format), used to voxelize samples to be processed by SparseNetwork.
\end{itemize}

\begin{table}
\centering{}
\begin{tabular}{|c|c|c|c|c|c|c|}
\hline
layer \#  & layer type & size  & stride & channels & spatial size & sparsity (\%)\footnotemark \tabularnewline
\hline
0 & Data input & - & - & 1 & 126 & 0.18 \tabularnewline
1 & Sparse Convolution & 2 & 1 & 8 & 125 & - \tabularnewline
2 & Leaky ReLU ($\alpha$ = 0.33) & - & - & 32 & 125 & 0.35 \tabularnewline
3 & Sparse MaxPool & 3 & 2 & 32 & 62 & 0.69 \tabularnewline
4 & Sparse Convolution & 2 & 1 & 256 & 61 & - \tabularnewline
5 & Leaky ReLU ($\alpha$ = 0.33) & - & - & 64 & 61 & 1.07 \tabularnewline
6 & Sparse MaxPool & 3 & 2 & 64 & 30 & 1.93 \tabularnewline
7 & Sparse Convolution & 2 & 1 & 512 & 29 & - \tabularnewline
8 & Leaky ReLU ($\alpha$ = 0.33) & - & - & 96 & 29 & 3.26 \tabularnewline
9 & Sparse MaxPool & 3 & 2 & 96 & 14 & 7.32 \tabularnewline
10 & Sparse Convolution & 2 & 1 & 768 & 13 & - \tabularnewline
11 & Leaky ReLU ($\alpha$ = 0.33) & - & - & 128 & 13 & 15.14 \tabularnewline
12 & Sparse MaxPool & 3 & 2 & 128 & 6 & 46.30 \tabularnewline
13 & Sparse Convolution & 2 & 1 & 1024 & 5 & - \tabularnewline
14 & Leaky ReLU ($\alpha$ = 0.33) & - & - & 160 & 5 & 97.54 \tabularnewline
15 & Sparse MaxPool & 3 & 2 & 160 & 2 & 100.00 \tabularnewline
16 & Sparse Convolution & 2 & 1 & 1280 & 1 & - \tabularnewline
17 & Leaky ReLU ($\alpha$ = 0.33) & - & - & 192 & 1 & 100.00 \tabularnewline
\hline
\end{tabular}
\caption{S3DCNN Network architecture.}
\label{tab:net-architecture}
\end{table}

\footnotetext{Last column ``sparsity'' is computed for render size = $40$ and averaged for all samples}

\section{Experiments}
\label{sec:4}
\subsection{ModelNet40 dataset}
In our experiments we used well known data set of 3D objects ModelNet40.
It is a subset of 40 classes of larger data set called ModelNet \cite{wu20153d} that contains different 3D CAD models in OFF format.

\begin{figure}
	\centering
    \includegraphics[width=\textwidth]{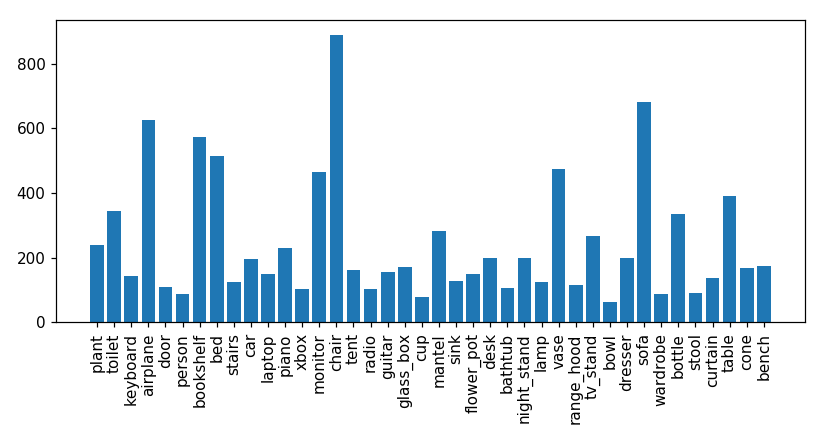}
    \caption{ModelNet40 data set: distribution of samples per class.}
    \label{fig:modelnet_classes}
\end{figure}

The total size of ModelNet40 data set $12311$.
The data set is split into training and test subsets, their sizes are $9843$ and $2468$ correspondingly.
The data set is not balanced.
Number of samples per class vary: from 64 to 889, see Figure \ref{fig:modelnet_classes}.

\subsection{Implementation details}
To demonstrate the impact that the triplet based training has on the performance of CNN descriptors
we use a deep network architecture shown in a Table~\ref{tab:net-architecture}. This network was implemented in PySparseConvNet, which is our modification of the SparseConvNet library \cite{graham2014spatially}. Besides new loss functions PySparseConvNet can be accessed from Python for a more interactive usage.

When forming a triplet for training we choose uniformly randomly a positive pair of objects from one class
and select a negative sample uniformly randomly from one of other classes.

For the optimization we use the SGD \cite{bottou-tricks-2012}, and the training is done
\begin{itemize}
\item in batches of size from $45$ to $90$ depending on a GPU video memory,
\item with a learning rate of $0.002$,
\item and a momentum equal to $0.99$.
\end{itemize}
Training can take up to a week on a server with advanced GPU, such as NVIDIA Titan X or GTX980ti.

\begin{figure}
\centering
\begin{tabular}{ccccc}
  \includegraphics[width=0.19\columnwidth]{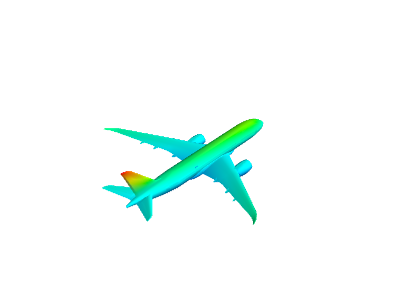} &
  \includegraphics[width=0.19\columnwidth]{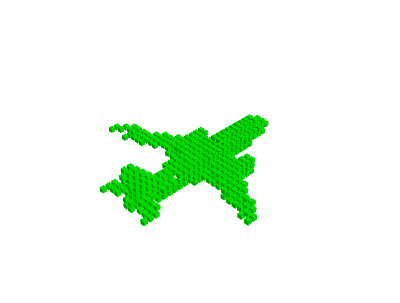} &
  \includegraphics[width=0.19\columnwidth]{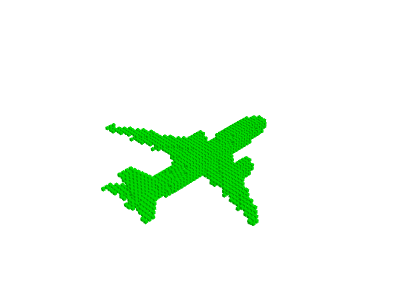} &
  \includegraphics[width=0.19\columnwidth]{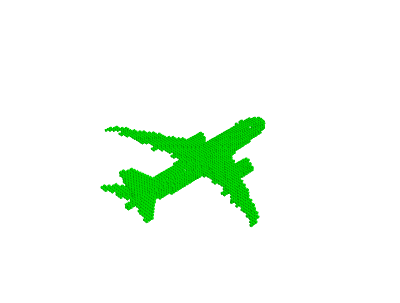} &
  \includegraphics[width=0.19\columnwidth]{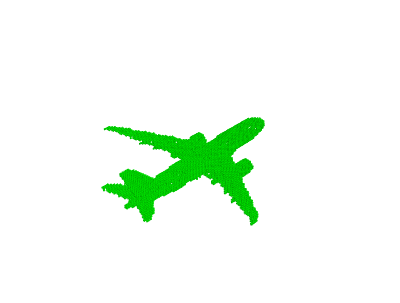} \\
% (a) first & (b) second \\
 \includegraphics[width=0.19\columnwidth]{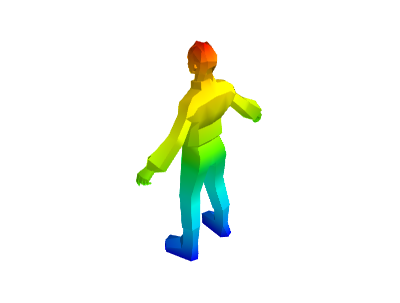} &
 \includegraphics[width=0.19\columnwidth]{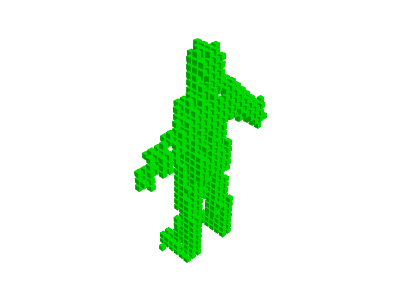} &
 \includegraphics[width=0.19\columnwidth]{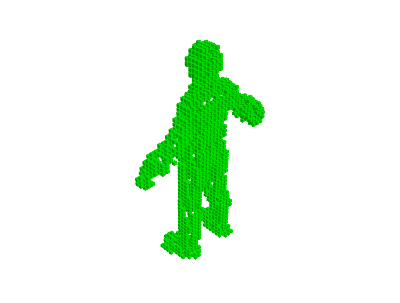} &
 \includegraphics[width=0.19\columnwidth]{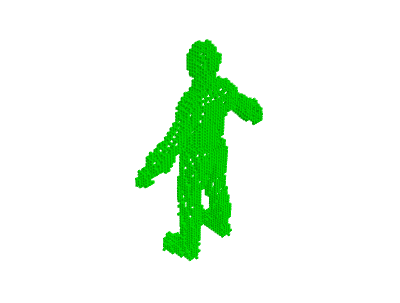} &
 \includegraphics[width=0.19\columnwidth]{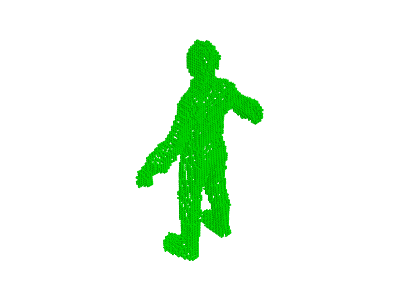} \\

 \includegraphics[width=0.19\columnwidth]{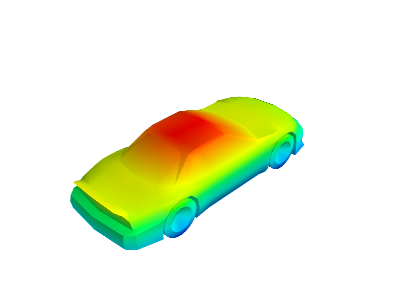} &
 \includegraphics[width=0.19\columnwidth]{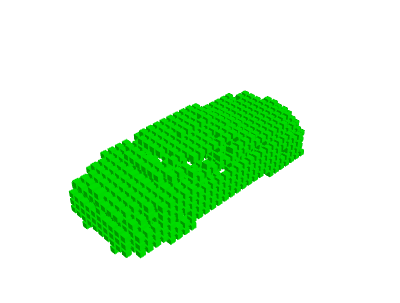} &
 \includegraphics[width=0.19\columnwidth]{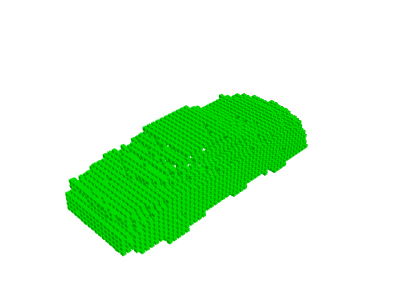} &
 \includegraphics[width=0.19\columnwidth]{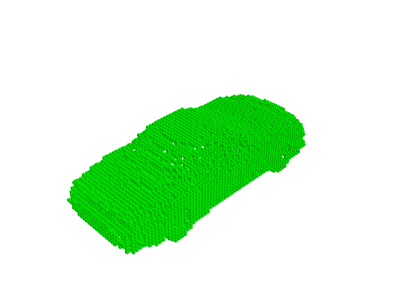} &
 \includegraphics[width=0.19\columnwidth]{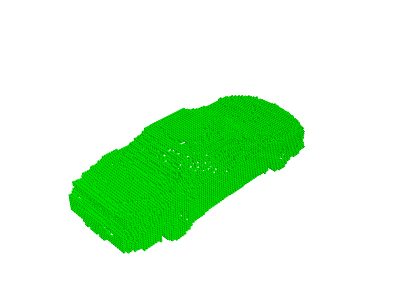} \\
\end{tabular}
\caption{Examples of some objects voxelizations at different resolutions $30$, $50$, $70$, $100$ (from left to right), left-most objects are depicted using original meshes}
\label{fig:voxels-examples}
\end{figure}

We train Sparse 3D Convolutional Neural Network (S3DCNN) on the 3D shape classification dataset by splitting it into  training and validation subsets, adding augmentation of data to achieve rotational and translational invariance. After training a model on a dataset of pairs, we use it to embed voxel representations of 3D meshes into $192$-dimensional space. The retrieval consist of ranking search objects by a cosine distance of vectors from a query vector.

The most popular metrics for evaluating retrieval performance are
\begin{itemize}
\item Precision-Recall Curve shows a trade-off between these two measures and how quickly the precision drops with the recall increase,
\item Mean average precision (mAP). Given a query, its average precision is the average of all precision values computed on all relevant objects in the retrieved list. Given several queries, the mean average precision (mAP) is the mean of average precisions for these queries. 
\end{itemize}
We evaluated mAP for different voxel rendering sizes of 3D shapes both at train and test times, see also Figure~\ref{fig:voxels-examples}.

To check if our model is comparable with other architectures, we consider a classification task. So, we trained our model for the classification task using the ModelNet40 train subset with 
\begin{itemize}
\item SoftMax last layer for $200$ epochs,
\item with exponentially discounting learning rate,
\item and performed retrieval evaluation on the test subset,
\item taking $20$ images from every class, and ranking them w.r.t their $L2$-norm by activations taken from the $17$-th layer.
\end{itemize}

Results of these experiments are provided in Table~\ref{tab:classification}. We can see that in case of classification task setup our model is comparable in terms of the classification accuracy, but mAP values are worse. But in case of metric learning performace of S3DCNN on mAP metric is much better.
Superior performance of retrieval task with MVCNN is not a surprising result, since MVCNN uses neural nets, pre-trained on ImageNet. On the other hand our model only requires 3D Shape dataset to learn.

\begin{figure}[!tbp]
\vspace{-20pt}
\centering
\begin{minipage}[b]{0.45\textwidth}
  \centering
  \includegraphics[width=\columnwidth]{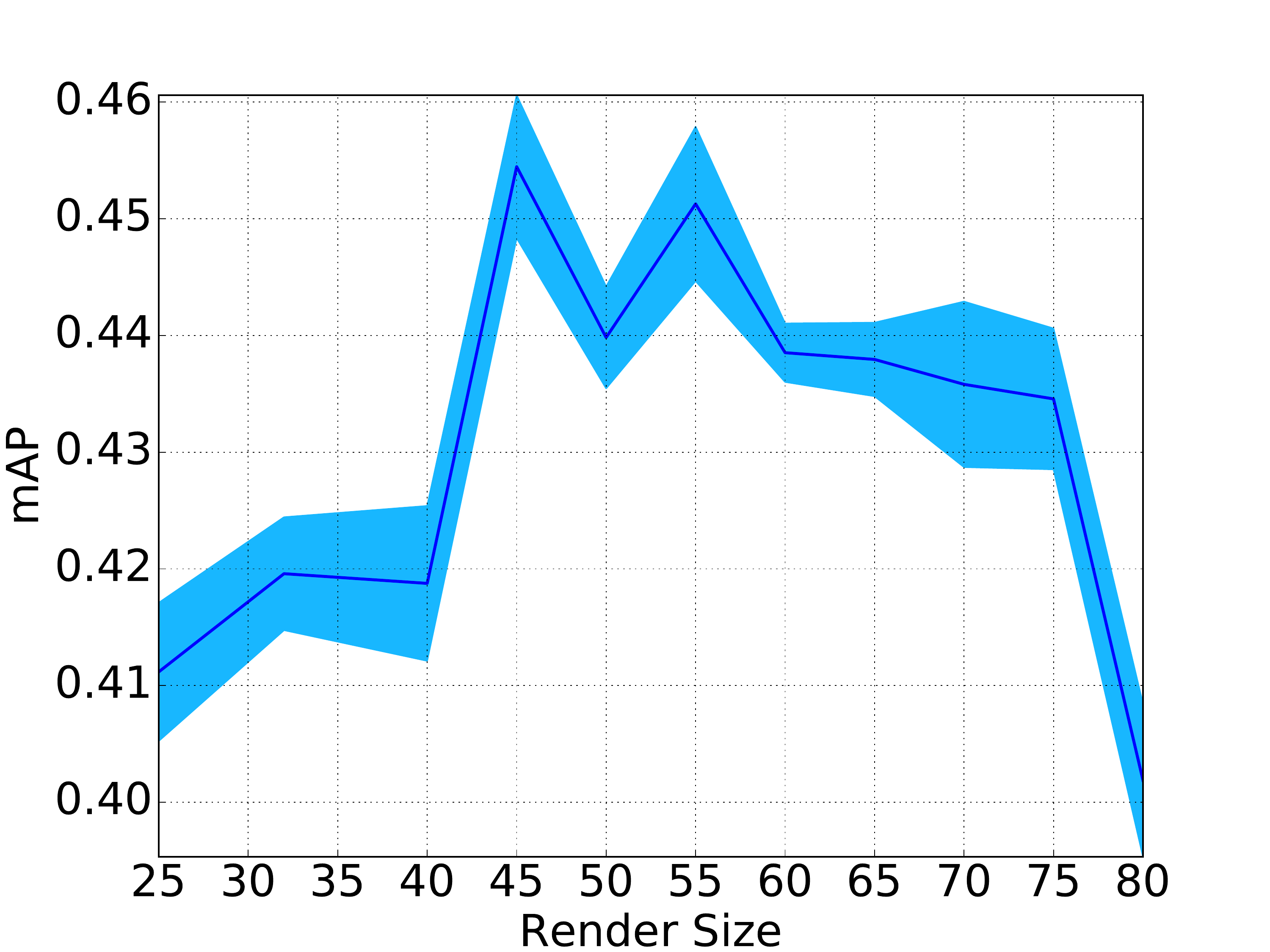}
  \caption{Dependence of the retrieval performance on the input spatial resolution}
  \label{fig:map_for_rs}
\end{minipage}
\begin{minipage}[b]{0.45\textwidth}
  \centering
  \includegraphics[width=\columnwidth]{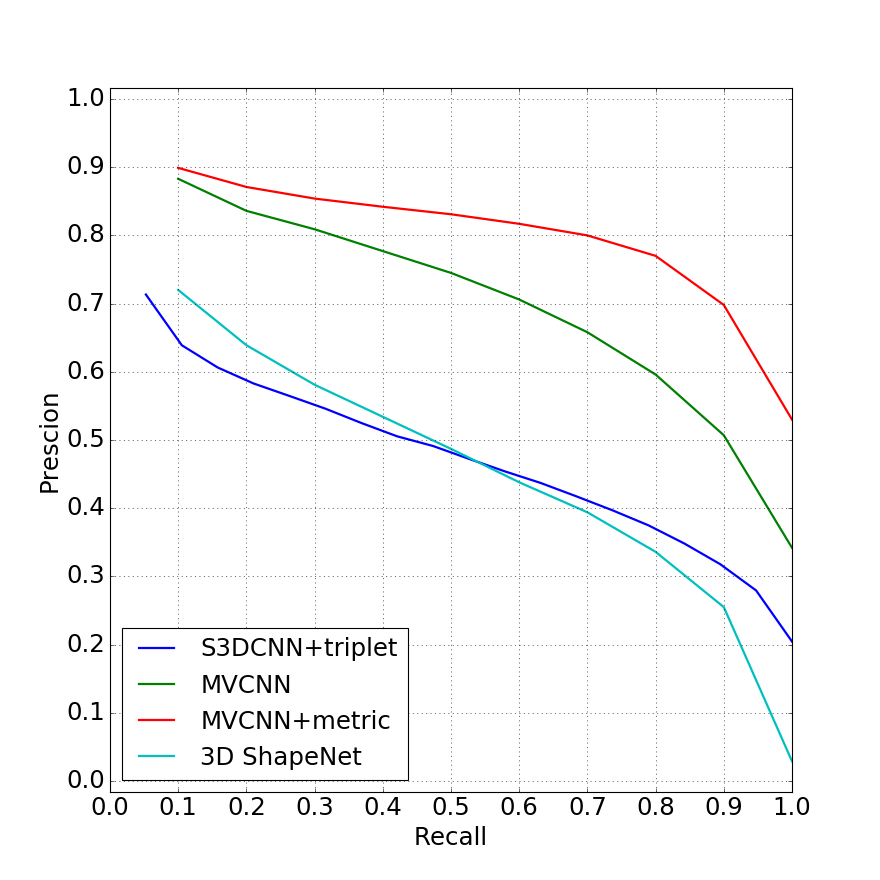}
  \caption{Precision-Recall curve for our method}
  \label{fig:pr_curve}
\end{minipage}
\end{figure}

In Figure~\ref{fig:map_for_rs} we provide the dependence of mAP on the input spatial resolution. We can see that the retrieval performance improves with increase in the input spatial resolution up to around $45-50$, after that it drops slightly and goes to plateau. It can be attributed to the insufficient amount of layers for the same scale of features, that can be separated in higher layers. Light blue color shows range of mAP on validation for top $30$ trained architectures.

We would like to note that in Figure~\ref{fig:map_for_rs} mAP values provided for different validation epochs and variability of best model can be explained by difference in total learning time.

% \begin{itemize}
% \item In Table~\ref{tab:classification} for the retrieval we used features from the last but one layer of the network,
% \item In Figure~\ref{fig:map_for_rs} we used learning with the triplet loss, for which we still have to better adjust the architecture and learning rate schedule.
% \end{itemize}

\section{Results}
\label{sec:5}
We found that the retrieval performance improves with increase in the input spatial resolution. However, such an effect is difficult to check experimentally and to use in practice, as e.g. for usual 3D dense CNNs the computational time is prohibitively large. In our case, data sparsity helps us to process data in reasonable time even with input resolution up to $100^3$ voxels, therefore we can benefit from the increase of the input spatial resolution when performing retrieval.
In Figure~\ref{fig:pr_curve} we can see that our method is comparable to \cite{wu20153d} in low recall, and better at higher recall values, that indicates better scalability of our method.
In Table~\ref{tab:classification} for the retrieval we used features from the one before last layer of the network of size 192, which in  comparison to 4000 in 3DShapeNet model \cite{wu20153d} is 20 times smaller but achieves almost the same retrieval metrics.

We evaluated our network architecture described in Table~\ref{tab:net-architecture} on popular state-of-the-art frameworks for Deep Learning, such as Tensorflow\cite{tensorflow2015-whitepaper} on GPU and Theano\cite{2016arXiv160502688short} on CPU.
Using Keras\cite{chollet2015keras} 2.0.2 with Tensorflow\cite{tensorflow2015-whitepaper} 1.2.1 backend on Nvidia Titan X GPU with 12Gb of GPU memory, we were able to exhaust all of it with batch size equal to 12, and performed forward passes on average 0.0301 seconds/sample, which is comparable to processing speed of our implementation with render size of about 60-70.
Other setup was an implementation of our network architecture on Keras with Theano backend using Intel i7-5820K 6-core CPU processor, took 1.53 seconds/sample, which is significantly slower.
% We provide training code for all experiments in our repository\footnote{\url{https://github.com/gangiman/PySparseConvNet}}.

\begin{table}[t]
  \caption{Evaluation on Modelnet40}
  \label{tab:classification}
  \centering
  \begin{tabular}{llll}
    \toprule
    method & Classification & Retrieval AUC & Retrieval mAP \\
    \midrule
    3DShapeNet \cite{wu20153d} & 77.32\% & 49.94\% & 49.23\% \\
    MVCNN \cite{su15mvcnn} & 90.10\% & --- & 80.20\% \\
    VoxNet \cite{maturana2015voxnet} & 83.00\% & --- & --- \\
    VRN \cite{brock2016generative} & 91.33\% & --- & --- \\
    \textbf{S3DCNN (proposed)} & \textbf{90.30}\% & \textbf{36.05}\% & \textbf{33.67}\% \\
    \textbf{S3DCNN + triplet (proposed)} & --- & \textbf{48.81}\% & \textbf{46.71}\% \\
    \bottomrule
  \end{tabular}
\end{table}

\subsubsection*{Acknowledgments}
We are very grateful to Dmitry Yarotsky for his contribution to this research project. Big Thanks to Benjamin Graham for some useful comments and ideas. Thanks to Rasim Akhunzyanov for his help in debugging the PySparseConvNet code.

The research was partially supported by the Russian Science Foundation grant (project 14-50-00150).

\clearpage
\bibliographystyle{abbrv}
\bibliography{bibliography}

\end{document}